\def\year{2019}\relax
\documentclass[letterpaper]{article} 
\usepackage{aaai19}  
\usepackage{times}  
\usepackage{helvet}  
\usepackage{courier}  
\usepackage{url}  
\usepackage{multirow}
\usepackage{amsmath}
\usepackage{amssymb}
\usepackage{graphicx,dblfloatfix}  
\usepackage{algorithmicx}
\usepackage{algorithm}
\usepackage{algpseudocode}
\begin{document}
%
\title{Data Fine-tuning}
\author{Saheb Chhabra, Puspita Majumdar, Mayank Vatsa, Richa Singh\\
IIIT-Delhi, India\\
{\{sahebc, pushpitam, mayank, rsingh\}@iiitd.ac.in}\\
}
\maketitle
\begin{abstract}
In real-world applications, commercial off-the-shelf systems are utilized for performing automated facial analysis including face recognition, emotion recognition, and attribute prediction. However, a majority of these commercial systems act as black boxes due to the inaccessibility of the model parameters which makes it challenging to fine-tune the models for specific applications. Stimulated by the advances in adversarial perturbations, this research proposes the concept of Data Fine-tuning to improve the classification accuracy of a given model without changing the parameters of the model. This is accomplished by modeling it as data (image) perturbation problem. A small amount of ``noise'' is added to the input with the objective of minimizing the classification loss without affecting the (visual) appearance. Experiments performed on three publicly available datasets LFW, CelebA, and MUCT, demonstrate the effectiveness of the proposed concept.
\end{abstract}

\section{Introduction}
With the advancements in machine learning (specifically deep learning), ready to use Commercial Off-The-Shelf (COTS) systems are available for automated face analysis, such as face recognition \cite{ding2017trunk}, emotion recognition \cite{fan2016video}, and attribute prediction \cite{hand2018doing}. However, often times the details of the model are not released which makes it difficult to update it for any other task or datasets. This renders the model's effectiveness as a black-box model only. To illustrate this, let $\mathbf{X}$ be the input data for a model with weights $\mathbf{W}$ and bias $b$. This model can be expressed as:

\begin{eqnarray}
\phi(\mathbf{W}\mathbf{X} + b)
\end{eqnarray}
If the source of the model is available, model fine-tuning is used to update the parameters. However, as mentioned above, in black box scenarios, the model parameters, $\mathbf{W}$ and $b$ cannot be modified, as the user does not have access to the model.

\begin{figure}[t]
\centering
\includegraphics[scale = 0.38]{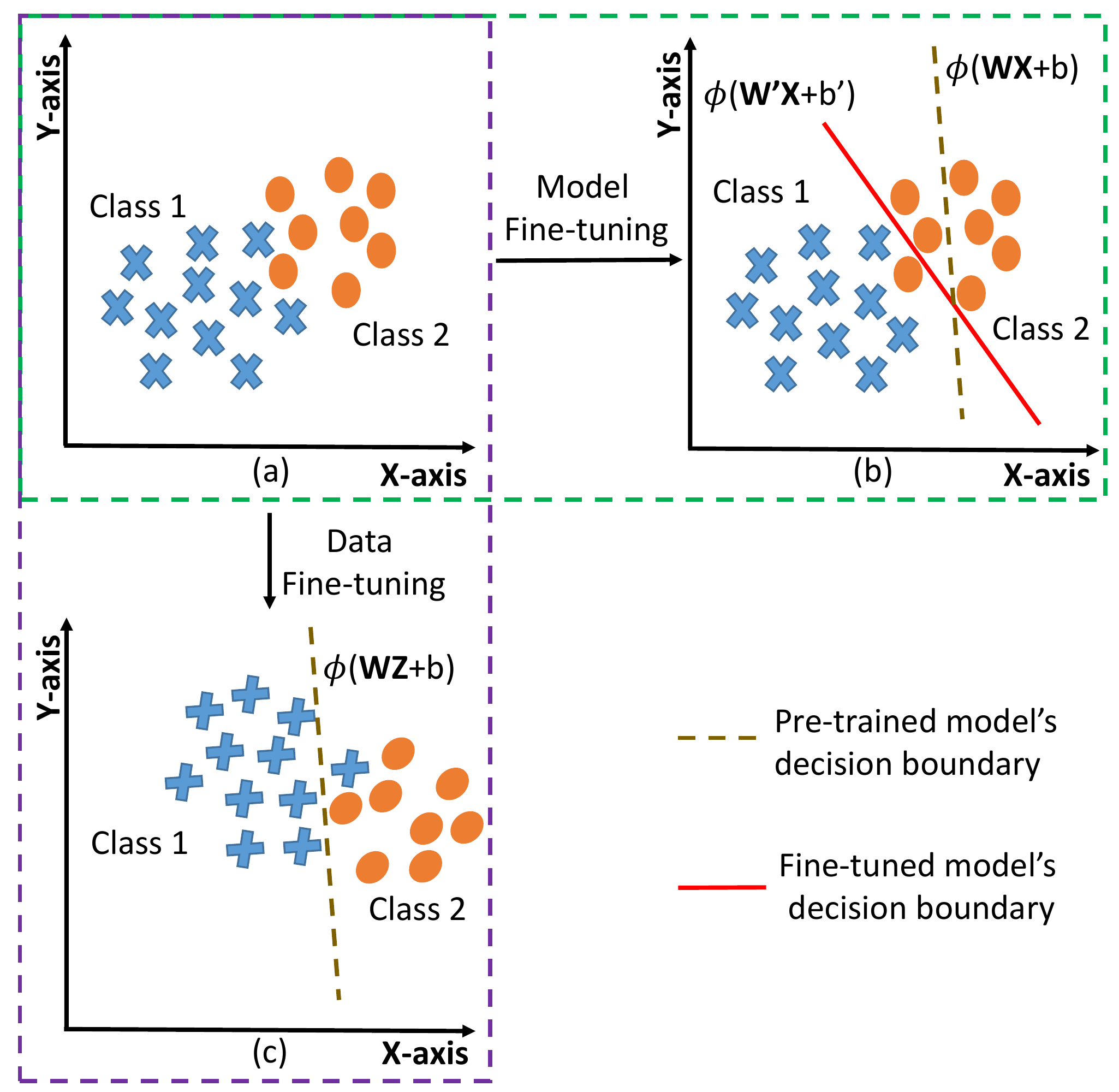}
\caption{Illustration of model fine-tuning and data fine-tuning: (a) represents the data distribution with two classes. (b) represents Model Fine-tuning where the model's decision boundary shifts corresponding to the input data, and (c) represents Data Fine-tuning where the input data shifts corresponding to model's decision boundary (best viewed in color).}
\label{fig:Visual}
\end{figure}

\begin{figure*}
\centering
\includegraphics[scale = 0.29]{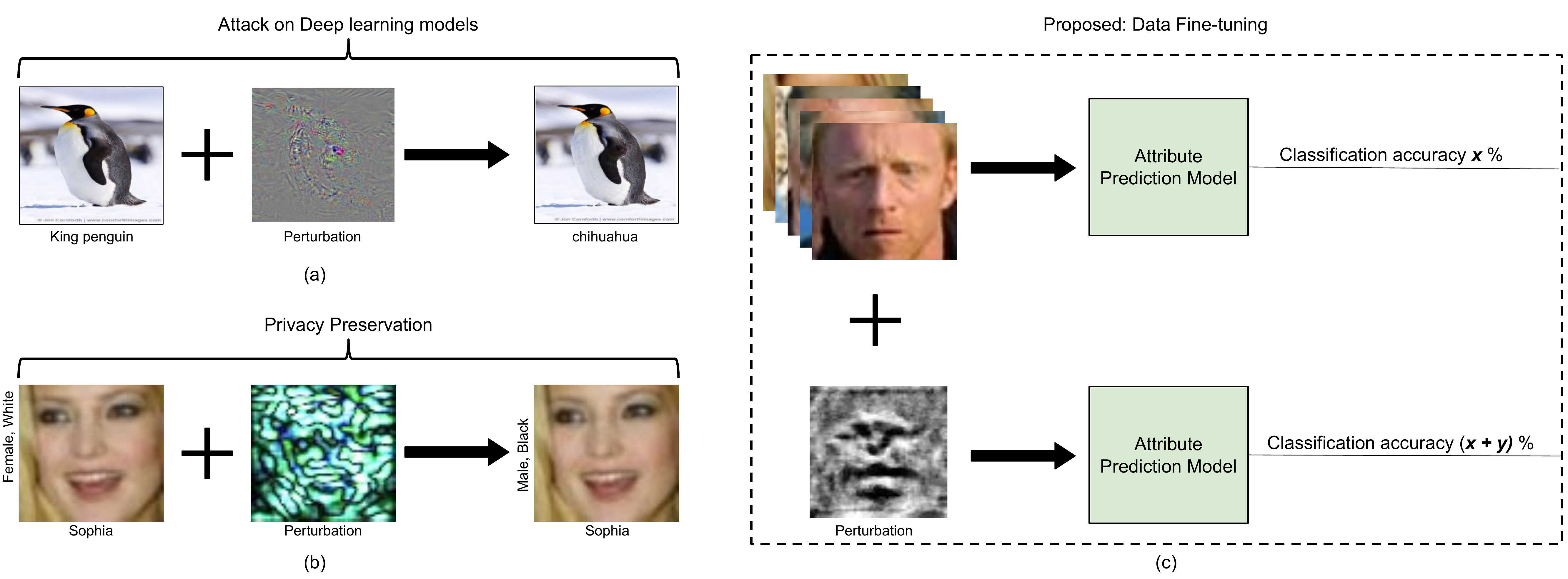}
\caption{Comparing the concept of adversarial perturbation with data fine-tuning. (a) Adversarial perturbation: shows the application of perturbation in attacking deep learning models \cite{xie2017mitigating}. (b) Privacy preservation: perturbation can be used to anonymize the attributes by preserving the identity of the input image \cite{chhabra2018anonymizing}. (c) Data Fine-tuning:  illustrates the proposed application of perturbation in enhancing the performance of a model (best viewed in color).}
\label{fig:Related_Work}
\vspace{-10pt}
\end{figure*}

``\textit{Can we enhance the performance of a black-box system for a given dataset?}'' To answer this question, in this research, we present a novel concept termed as \textbf{Data Fine-tuning (DFT)}, wherein the input data is adjusted corresponding to the model's unseen decision boundary. To the best of our knowledge, this is the first work towards data fine-tuning to enhance the performance of a given black box system. As shown in Figure \ref{fig:Visual}, the proposed data fine-tuning adjusts the input data $\mathbf{X}$ whereas, in the model fine-tuning approach (MFT), the parameters ($\mathbf{W}$, $b$) are adjusted for optimal classification. 

Mathematically, model fine-tuning is:
\begin{eqnarray}
\phi(\mathbf{WX} + b) \xrightarrow[\text{}]{\text{MFT}} \phi(\mathbf{W'X} + b')
\end{eqnarray}
and data fine-tuning can be written as:
\begin{eqnarray}
\phi(\mathbf{WX} + b) \xrightarrow[\text{}]{\text{DFT}} \phi(\mathbf{WZ} + b)
\end{eqnarray}
where, MFT and DFT are model fine-tuning\footnote{Various data augmentation techniques have also been used for model fine-tuning \cite{salamon2017deep,um2017data,wu2018conditional}} and data fine-tuning, respectively. ($\mathbf{W'}$,$b'$) are the parameters after MFT and $\mathbf{Z}$ is the perturbed version of input $\mathbf{X}$ after data fine-tuning. 

In this research, the proposed data fine-tuning is achieved using adversarial perturbation. For this purpose, samples in the training data are uniformly perturbed and the model is trained iteratively on this perturbed training data to minimize classification loss. After each iteration, optimization is performed over the perturbation noise and added to the training data. At the end of the training, a single uniform perturbation is learned corresponding to a dataset.  As a case study, the proposed algorithm is evaluated for facial attribute classification. It learns a single universal perturbation for a given dataset to improve facial attribute classification while preserving the visual appearance of the images. Experiments are performed on three publicly available datasets and results showcase enhanced performance of black box systems using data fine-tuning. 
\subsection{Related Work}

In the literature, perturbation is studied from two perspectives: (i) privacy preservation and (ii) attacks on deep learning models. For privacy preservation, several techniques utilizing data perturbation are proposed. \cite{jain2011min} proposed min max normalization method to perturb data before using in data mining applications. \cite{last2014improving} proposed a data publishing method using NSVDist. Using this method, the sensitive attributes of the data are published as the frequency distributions. Recently, \cite{chhabra2018anonymizing} proposed an algorithm to anonymize multiple facial attributes in an input image while preserving the identity using adversarial perturbation. \cite{li2018secure} proposed Random Linear Transformation with Condensed Information-Support Vector Machine to convert the condensed information to another random vector space to achieve safe and efficient data classification. 

\cite{szegedy2013intriguing} demonstrated that application of imperceptible perturbation could lead to the misclassification of an image. \cite{papernot2016limitations} created an adversarial attack by restricting $l_0$-norm of the perturbation where only a few pixels of an image are modified to fool the classifier. \cite{carlini2017towards} introduced three adversarial attacks and showed the failure of defensive distillation \cite{carlini2016defensive} for targeted networks. By adding perturbation, \cite{kurakin2016adversarial} replaced the original label of the image with the label of least likely predicted class by the classifier. This lead to the poor classification accuracy of Inception v3. \cite{su2017one} proposed a one-pixel attack in which three networks are fooled by changing one pixel per image. Universal adversarial perturbation proposed by \cite{moosavi2017universal} can fool a network when applied to any image. This overcomes the limitation of computing perturbation on every image. \cite{goswami2018unravelling} proposed a technique for automatic detection of adversarial attacks by using the abnormal filter response from the hidden layer of the deep neural network. Further, a novel technique of selective dropout is proposed to mitigate the adversarial attacks. \cite{goel2018smartbox} developed SmartBox toolbox for detection and mitigation of adversarial attacks against face recognition.

Existing literature demonstrates the application of adversarial perturbation for performing attacks on deep learning models and in privacy preservation (Figure \ref{fig:Related_Work}(a) and (b)). However, data fine-tuning using adversarial perturbation (Figure \ref{fig:Related_Work}(c)) for enhancing the performance of a model is not yet explored.

\begin{figure}[t]
\centering
\includegraphics[scale = 0.45]{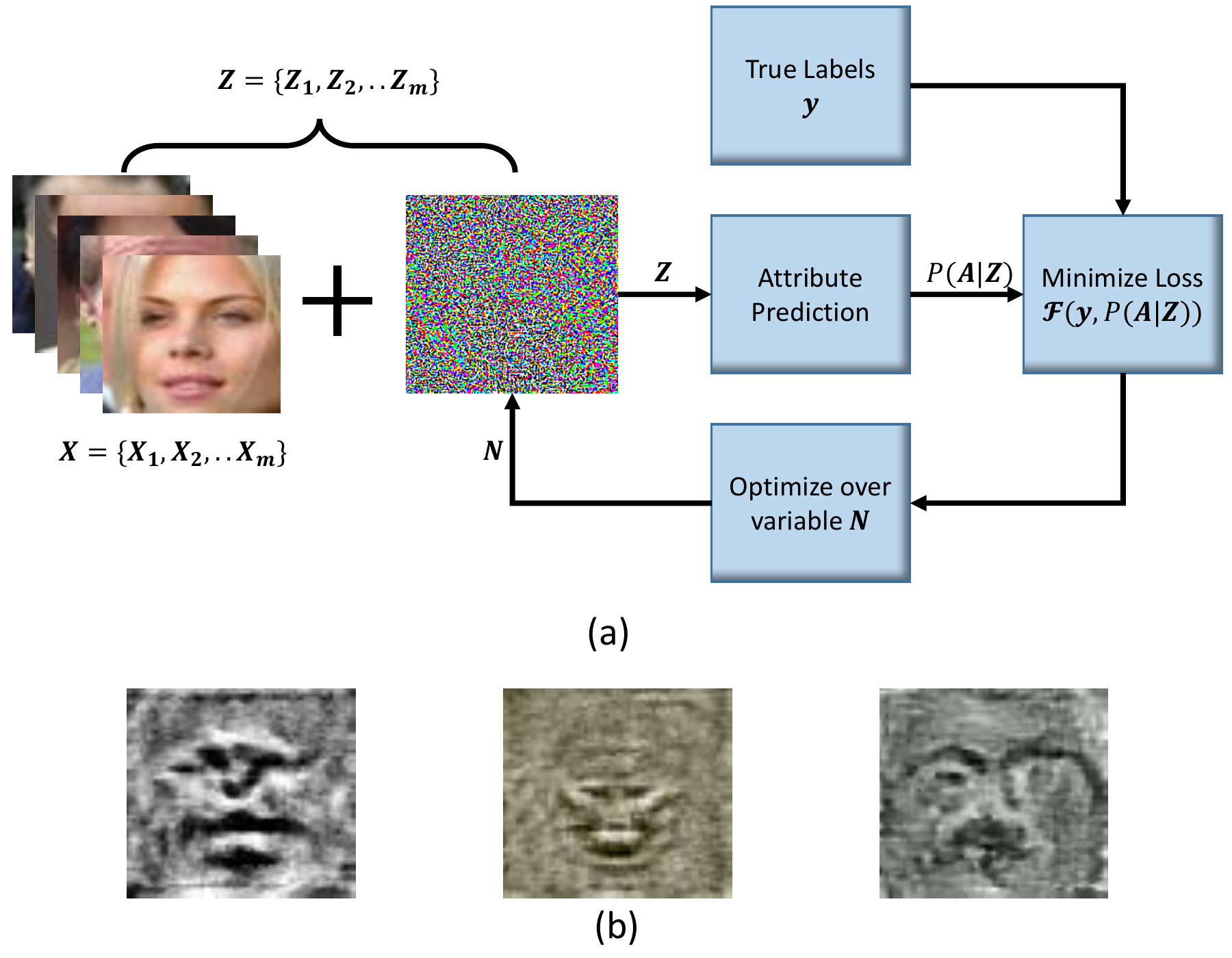}
\caption{(a) Block diagram illustrating the steps of the proposed algorithm. In the first step, perturbation is initialized with zero image and added to the original training data. In the next step, perturbed training data is given as input to the (attribute prediction) model followed by the computation of loss. After that, optimization is performed over perturbation and added to the training data. (b) Some samples of the learned perturbation using the proposed algorithm. The first two visualizations correspond to the perturbation learned for `Smiling' attribute of LFW and CelebA datasets, respectively. The third visualization corresponds to the `Gender' attribute of the MUCT dataset (best viewed in color).}
\label{fig:Block_Diagram}
\end{figure}

\section{Proposed Approach: Data Fine-tuning}
Considering a black-box system as a pre-trained model, the problem statement can be defined as ``given the dataset $\mathbf{D}$ and pre-trained model $M$, learn a perturbation vector $\mathbf{N}$ such that adding noise $\mathbf{N}$ to $\mathbf{D}$ improves the performance of the model $M$ on $\mathbf{D}$''. There are two important considerations while performing data fine-tuning:
\begin{enumerate}
\item To learn a single universal perturbation noise for a given dataset.
\item The visual appearance of the image should be preserved after performing data fine-tuning. 
\end{enumerate}

The block diagram illustrating the steps involved in the proposed algorithm is shown in Figure \ref{fig:Block_Diagram}. The optimization process for data fine-tuning using adversarial perturbation with applications to facial attribute classification is discussed below. This same approach can be extended for other classification models.

Given the original training set $\mathbf{X}$ with $m$ number of images where each image, $\mathbf{X_k}$ has pixel values in the range $\{0, 1\}$, i.e., $\mathbf{X_k} \in [0, 1]$. Let $\mathbf{Z}$ be the perturbed training set generated by adding model specific perturbation noise $\mathbf{N}$ such that the pixel values of each output perturbed image $\mathbf{Z_k}$ ranges between $0$ to $1$, i.e., $\mathbf{Z_k} \in [0, 1]$. Mathematically, it is written as:


\begin{equation}
\mathbf{Z_k} = f(\mathbf{X_k} + \mathbf{N})
\end{equation}
$$\text{such that} \quad f(\mathbf{X_k} + \mathbf{N})  \in [0,1]$$
where, $f(.)$ represents the function to transform an image in the range of 0 to 1. In order to satisfy the above constraint, inspired by \cite{carlini2017towards}, the following function is used:


\begin{equation}
\label{eq:4}
\mathbf{Z_k} = \frac{1}{2}(tanh(\mathbf{X_k} + \mathbf{N})+1)
\end{equation}
For each image $\mathbf{X_k}$ there are $n$ number of attributes in the attribute set $\mathbf{A}$, where each attribute $\mathbf{A_i}$ has $C_j$ number of classes. For example, `Gender' attribute has two classes namely \{Male, Female\} while `Expression' attribute has three classes namely \{Happy, Sad, Anger\}. Mathematically, it is written as: 

\begin{equation}
\mathbf{A} = \{\mathbf{A_1}(C_1), \mathbf{A_2}(C_2),...\mathbf{A_n}(C_n)\} 
\end{equation}

The pre-trained attribute prediction model for attribute $\mathbf{A_i}$ is represented as $\phi_{A_i}(\mathbf{X_k}, \mathbf{W}, b)$, where $\mathbf{W}$ is the weight matrix and $b$ is the bias. The output attribute score of any image $\mathbf{X_k}$ is written as:

\begin{equation}
P(\mathbf{A_i}|\mathbf{X_k}) = \phi_{\mathbf{A_i}}(\mathbf{X_k}, \mathbf{W}, b)
\end{equation}
where, $P(\mathbf{A_i}|\mathbf{X_k})$ represents the output attribute score of the input image $\mathbf{X_k}$ for attribute $\mathbf{A_i}$. In order to perform data fine-tuning, perturbation $\mathbf{N}$ is added to each input image $\mathbf{X_k}$ to get the output perturbed image $\mathbf{Z_k}$ using Equation \ref{eq:4}. Here, $\mathbf{N}$ is the perturbation variable to be optimized. The output attribute score of the perturbed image $\mathbf{Z_k}$ is represented as:

\begin{equation}
P(\mathbf{A_i}|\mathbf{Z_k}) = \phi_{\mathbf{A_i}}(\mathbf{Z_k}, \mathbf{W}, b)
\end{equation}
 In order to enhance the model's performance for attribute $\mathbf{A_i}$, the distance between the true class and attribute predicted score of the perturbed image is minimized which is expressed as:

\begin{equation}
\label{eq:5}
\min_{\mathbf{N}} \quad \mathcal{F}(\mathbf{y_{i,k}}, P(\mathbf{A_i}|\mathbf{Z_k}))
\end{equation}
where, $\mathcal{F}(.,.)$ represents the function to minimize the distance between the true class and the predicted class.  $\mathbf{y_{i,k}}$ represents the true class of attribute $\mathbf{A_i}$ in one hot encoding form of the original image $\mathbf{X_k}$. To preserve the visual appearance of the output perturbed image $\mathbf{Z_k}$, the distance between original image $\mathbf{X_k}$ and the perturbed image $\mathbf{Z_k}$ is minimized. Thus, the above equation is updated as:

\begin{equation}
\min_{\mathbf{N}} \quad \mathcal{F}(\mathbf{y_{i,k}}, P(\mathbf{A_i}|\mathbf{Z_k})) + H(\mathbf{X_k}, \mathbf{Z_k})
\end{equation}
where, $H$ represents the distance metric to minimize the distance between $\mathbf{X_k}$ and $\mathbf{Z_k}$. In this research, Euclidean distance metric is used to preserve the visual appearance of the image. Therefore,
\begin{equation}
\min_{\mathbf{N}} \quad \mathcal{F}(\mathbf{y_{i,k}}, P(\mathbf{A_i}|\mathbf{Z_k})) + ||\mathbf{X_k} - \mathbf{Z_k}||_F^2
\end{equation}

Since the output class score ranges between 0 and 1, the objective function in Equation (\ref{eq:5}) is formulated as:
\begin{equation}
\mathcal{F}(\mathbf{y_{i}}, P(\mathbf{A_i}|\mathbf{Z})) = \frac{1}{m}  \sum_{k=1}^{m} max(0, 1 -  \mathbf{y_{i,k}^T} P(\mathbf{A_i}|\mathbf{Z_k}))
\end{equation}
where, $i \in \{1,...,n\}$, and the term $\mathbf{y_{i,k}^T}  P(\mathbf{A_i}|\mathbf{Z_k})$ outputs the attribute score of the true class. As the above function $\mathcal{F}(\mathbf{y_i},  P(\mathbf{A_i}|\mathbf{Z}))$ is to be minimized, the term $max(0, 1 - \mathbf{y_{i,k}^T}  P(\mathbf{A_i}|\mathbf{Z_k}))$ enforces the output attribute score of the true class of the perturbed image $\mathbf{Z_k}$ towards one. 

\begin{figure}[t]
\centering
\includegraphics[scale = 0.27]{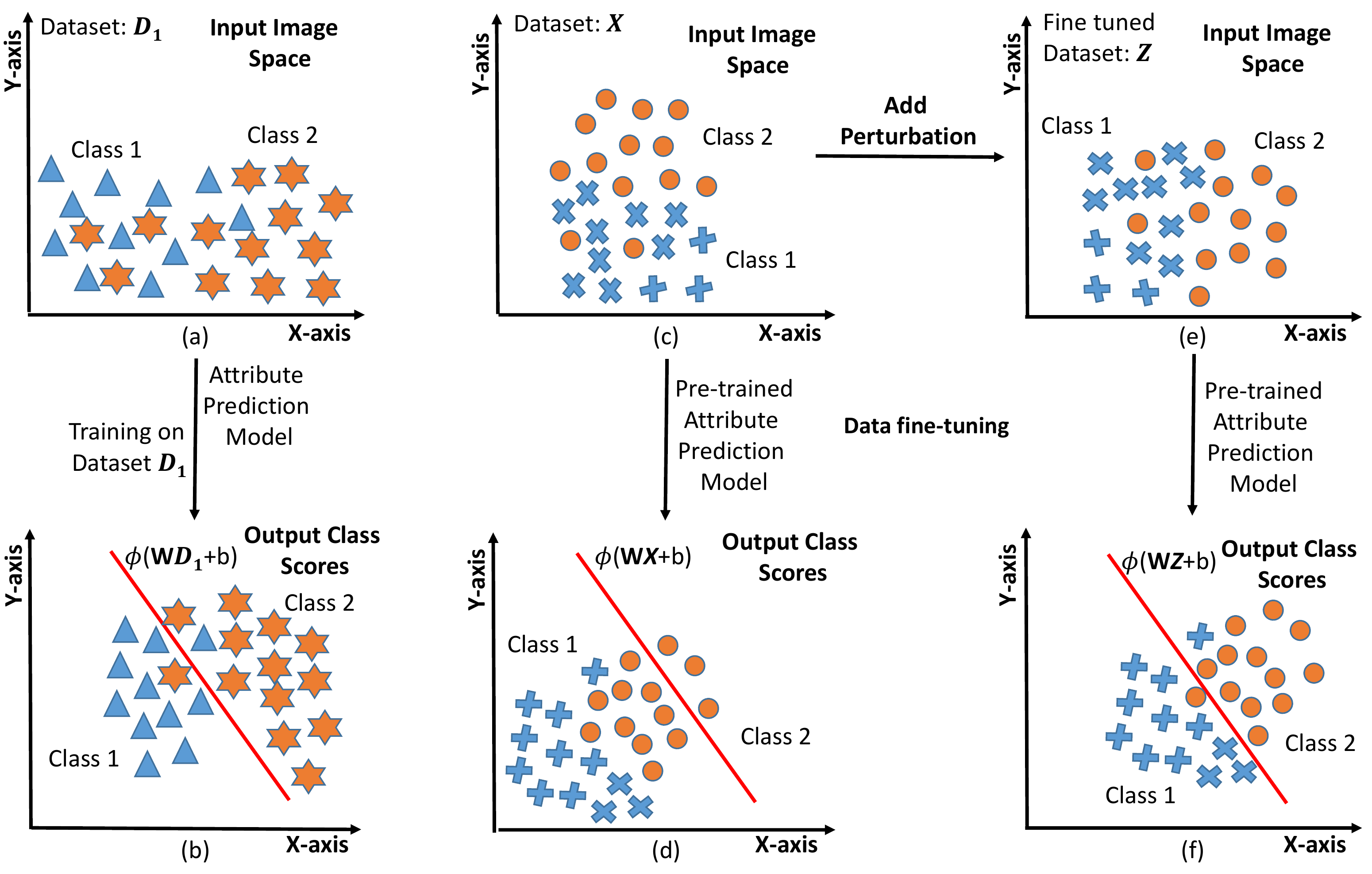}
\caption{Illustration of the proposed DFT algorithm. Figure (a)-(b) represents the training of attribute prediction model using dataset $\mathbf{D_1}$. (c)-(d) shows the performance of the trained attribute prediction model on dataset $\mathbf{X}$. (e)-(f) shows the performance of the fine-tuned dataset $\mathbf{Z}$ by adding perturbation on trained attribute prediction model. (Best viewed in color).}
\label{Proposed_Approach}
\end{figure}

Figure \ref{Proposed_Approach} illustrates the proposed algorithm with an example. Let $\mathbf{D_1}$ be the dataset with two classes in the input image space (Figure \ref{Proposed_Approach}(a)) and it is used to train a model, ${M_1}$. Model $M_1$ computes the decision boundary and projects the output class scores corresponding to the input data $\mathbf{D_1}$ as shown in Figure \ref{Proposed_Approach}(b). It is observed that the output class scores are well separated across the decision boundary for the dataset $\mathbf{D_1}$. Now, the pre-trained model $M_1$ is used for projecting the input dataset $\mathbf{X}$ (Figure \ref{Proposed_Approach}(c)). The decision boundary of the model $M_1$ remains fixed. The projected output class scores of the input data $\mathbf{X}$ are shown in Figure  \ref{Proposed_Approach}(d). It is observed that most of the data points of both the classes are projected on the same side of the decision boundary resulting in a high classification error. This is due to the change in the data distribution of the input dataset $\mathbf{X}$. To overcome this problem, input dataset $\mathbf{X}$ is fine-tuned by adding perturbation noise. Figure \ref{Proposed_Approach}(e) shows the fine-tuned dataset $\mathbf{Z}$ that is given as input to the model $M_1$. The projection of the fine-tuned dataset $\mathbf{Z}$ is shown in Figure \ref{Proposed_Approach}(f). On comparing the output class scores of the projection of input data $\mathbf{X}$ and fine-tuned data $\mathbf{Z}$, it is observed that several misclassified samples from $\mathbf{X}$ are correctly classified with the fine-tuned dataset $\mathbf{Z}$. 

\begin{table}[ht]
\centering
\scriptsize
\caption{Details of the experiments to show the efficacy of the proposed data fine-tuning for facial attribute classification.}
\label{Experimental_Details}
\begin{tabular}{|c|c|c|c|}
\hline
\multirow{2}{*}{\textbf{Experiment}}                                                                              & \textbf{}                                                                        & \textbf{Data Fine-tuning} & \multicolumn{1}{l|}{\textbf{Model Training}} \\ \cline{2-4} 
                                                                                                                  & Attribute                                                                        & Database                  & Database                                  \\ \hline
\multirow{5}{*}{\textbf{\begin{tabular}[c]{@{}c@{}}Black Box\\ Data \\Fine-tuning:\\ Intra Dataset\end{tabular}}}   & \multirow{3}{*}{Gender}                                                          & MUCT                      & MUCT                                      \\ \cline{3-4} 
                                                                                                                  &                                                                                  & LFW                       & LFW                                       \\ \cline{3-4} 
                                                                                                                  &                                                                                  & CelebA                    & CelebA                                    \\ \cline{2-4} 
                                                                                                                  & \begin{tabular}[c]{@{}c@{}}Smiling, Bushy \\ Eyebrows\\ Pale Skin\end{tabular}   & LFW                       & LFW                                       \\ \cline{2-4} 
                                                                                                                  & \begin{tabular}[c]{@{}c@{}}Smiling, Attractive, \\ Wearing Lipstick\end{tabular} & CelebA                    & CelebA                                    \\ \hline
\multirow{5}{*}{\textbf{\begin{tabular}[c]{@{}c@{}}Black Box \\ Data \\Fine-tuning: \\ Inter Dataset\end{tabular}}} & \multirow{3}{*}{Gender}                                                          & MUCT                      & LFW, CelebA                               \\ \cline{3-4} 
                                                                                                                  &                                                                                  & LFW                       & MUCT, CelebA                              \\ \cline{3-4} 
                                                                                                                  &                                                                                  & CelebA                    & MUCT, LFW                                 \\ \cline{2-4} 
                                                                                                                  & \begin{tabular}[c]{@{}c@{}}Smiling, Bushy\\ Eyebrows,\\ Pale Skin\end{tabular}   & LFW                       & CelebA                                    \\ \cline{2-4} 
                                                                                                                  & \begin{tabular}[c]{@{}c@{}}Smiling, Attractive,\\ Wearing Lipstick\end{tabular}  & CelebA                    & LFW                                       \\ \hline
\end{tabular}
\end{table}


\begin{figure*}[t]
\centering
\includegraphics[scale = 0.76]{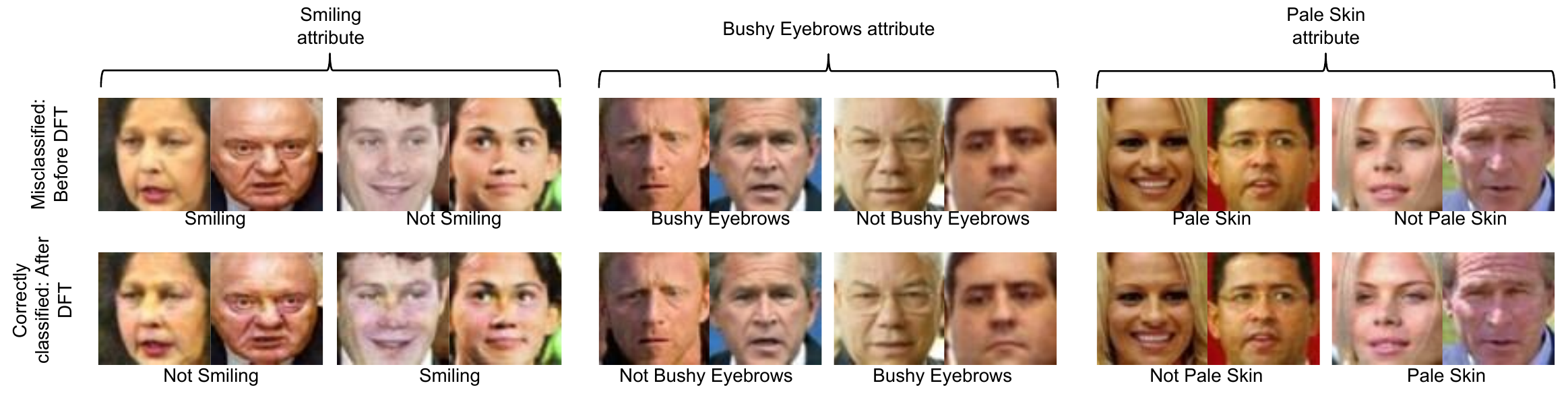}
\caption{Misclassified samples that are correctly classified after data fine-tuning. First row shows the images misclassified before data fine-tuning while the second row represents their correct classification after data fine-tuning. The first block of images correspond to the `Smiling' attribute, second block corresponds to `Bushy Eyebrows', while the third block corresponds to `Pale Skin' of the LFW dataset. (Best viewed in color).}
\label{fig:Mis_Correct_Classified}
\end{figure*}

\section{Datasets Protocol and Experimental Details}
The proposed algorithm is evaluated on three publicly available datasets for facial attribute classification: LFW \cite{huang2008labeled}, CelebA \cite{liu2015faceattributes}, and MUCT \cite{milborrow2010muct}. A comparison has also been performed between Data Fine-tuning and Model Fine-tuning. The details of each dataset and its protocol are described below : 


\textbf{LFW dataset} consists of 13,133 images of 5,749 subjects. Total 73 attributes are annotated with intensity values for each image. The attributes are binarized by considering positive intensity values as attribute present with label 1 and negative intensity values as attribute absent with label 0. The dataset is partitioned into 60\% training set, 20\% validation set, and 20\% testing set. 

\textbf{CelebA dataset} consists of 202,599 face images of more than 10,000 celebrities. For each image, 40 binary attributes are annotated such as Male, Smiling, and Bushy Eyebrows. Standard pre-defined protocol is followed for experiments and the dataset is partitioned into 162,770 images in the training set, 19,867 into validation set, and 19,962 images in the testing set.   

\textbf{MUCT dataset} consists of 3,755 images of 276 subjects out of which 131 are male and 146 are female. Viola-Jones face detector is applied on all the images, and the detector failed to detect 49 face images. Therefore, only 3,706 images are considered for further processing. These images are further partitioned into 60\% training set, 20\% validation set, and 20\% testing set corresponding to each class.

To evaluate the performance of data fine-tuning, two experiments are performed, (i) \textit{Black Box Data Fine-tuning: Intra Dataset} and (ii) \textit{Black Box Data Fine-tuning: Inter Dataset}. Both the experiments are performed on all the three datasets. Classification performance of the attributes is enhanced corresponding to the attribute classification model. To train the attribute classification model, pre-trained VGGFace \cite{parkhi2015deep} + NNET is used. Experimental details are also shown in Table \ref{Experimental_Details}.

\begin{table}[t]
\centering
\caption{Classification accuracy (in \%) of before and after Data Fine-tuning(DFT) for `Gender' attribute on LFW, CelebA, and MUCT datasets.}
\label{tab: Gender_Acc}
\begin{tabular}{|c|c|c|}
\hline
       & \textbf{Before DFT} & \textbf{After DFT} \\ \hline
LFW    & 87.94                & \textbf{91.17}               \\ \hline
CelebA & 82.13           & \textbf{83.08}          \\ \hline
MUCT   & 91.67           & \textbf{94.31}          \\ \hline
\end{tabular}
\end{table}

\begin{table}[t]
\centering
\scriptsize
\caption{Classification accuracy (in \%) before and after performing data fine-tuning for three attributes on the LFW and CelebA datasets.}
\label{tab: Single_Attr_Class}
\begin{tabular}{|c|c|c|c|c|c|c|}
\hline
\multirow{3}{*}{LFW}    & \multicolumn{2}{c|}{\textbf{Smiling}} & \multicolumn{2}{c|}{\textbf{Bushy Eyebrows}} & \multicolumn{2}{c|}{\textbf{Pale Skin}}        \\ \cline{2-7} 
                        & Before             & After            & Before                & After                & Before                  & After                \\ \cline{2-7} 
                        & 76.18              & \textbf{82.42}            & 68.34                 & \textbf{69.98}                & 72.83                   & \textbf{74.81}                \\ \hline
\multirow{3}{*}{CelebA} & \multicolumn{2}{c|}{\textbf{Smiling}} & \multicolumn{2}{c|}{\textbf{Attractive}}     & \multicolumn{2}{c|}{\textbf{Wearing Lipstick}} \\ \cline{2-7} 
                        & Before             & After            & Before                & After                & Before                  & After                \\ \cline{2-7} 
                        & 67.82              & \textbf{71.30}            & 70.48                 & \textbf{70.54}                & 80.95                   & \textbf{81.29}                \\ \hline
\end{tabular}
\end{table}

\begin{figure}[t]
\centering
\includegraphics[scale = 0.42]{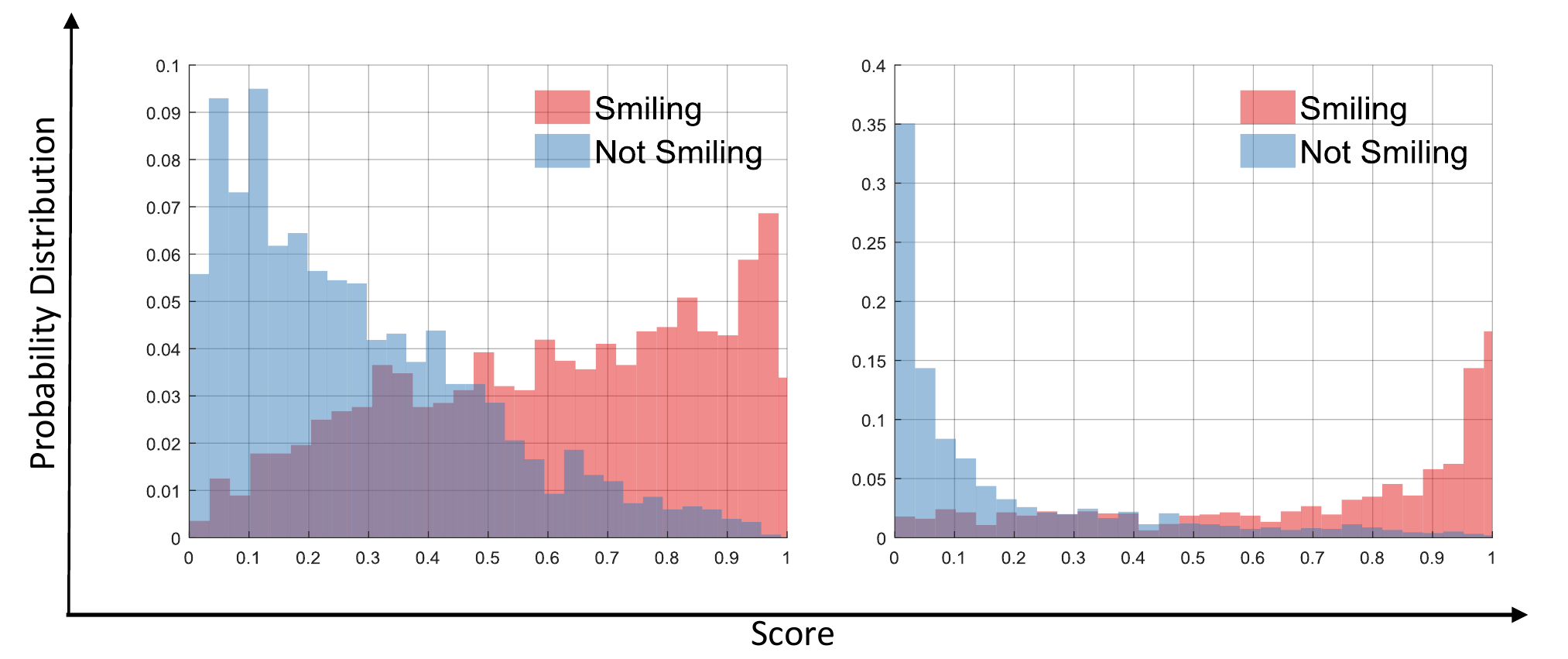}
\caption{Smiling attribute score distribution pertaining to before and after performing data fine-tuning on the LFW dataset. The left graph represents the score distribution before data fine-tuning and right graph represents the score distribution after data fine-tuning. (Best viewed in color).}
\label{fig: Score_Distr}
\end{figure}

\begin{table*}
\scriptsize
\centering
\caption{Confusion matrix of the LFW dataset for three attributes: `Smiling', `Bushy Eyebrows', `Pale Skin'.}
\label{Conf_mat}
\begin{tabular}{|c|c|c|c|c|c||c|c|c||c|c|c|}
\hline
\multicolumn{3}{|c|}{}                                                                                                                                                                                          & \textbf{\begin{tabular}[c]{@{}c@{}}Attribute\\ Class\end{tabular}} & \multicolumn{2}{c||}{\textbf{Prediction}}                                & \textbf{\begin{tabular}[c]{@{}c@{}}Attribute\\ Class\end{tabular}} & \multicolumn{2}{c||}{\textbf{Prediction}}                                                                                  & \textbf{\begin{tabular}[c]{@{}c@{}}Attribute\\ Class\end{tabular}} & \multicolumn{2}{c|}{\textbf{Prediction}}                                  \\ \hline
                              &                                                                                  &                                                                                              &                                                                    & Smiling        & \begin{tabular}[c]{@{}c@{}}Not \\ Smiling\end{tabular} &                                                                    & \begin{tabular}[c]{@{}c@{}}Bushy \\ Eyebrows\end{tabular} & \begin{tabular}[c]{@{}c@{}}Not Bushy \\ Eyebrows\end{tabular} &                                                                    & Pale Skin      & \begin{tabular}[c]{@{}c@{}}Not Pale \\ Skin\end{tabular} \\ \hline
\multirow{4}{*}{\textbf{LFW}} & \multirow{4}{*}{\textbf{\begin{tabular}[c]{@{}c@{}}Ground\\ Truth\end{tabular}}} & \multirow{2}{*}{\textbf{\begin{tabular}[c]{@{}c@{}}Before \\ Data Fine-tuning\end{tabular}}} & Smiling                                                            & 65.50          & 34.50                                                  & \begin{tabular}[c]{@{}c@{}}Bushy \\ Eyebrows\end{tabular}          & 77.17                                                     & 22.83                                                         & Pale Skin                                                          & 74.48          & 25.52                                                    \\ \cline{4-12} 
                              &                                                                                  &                                                                                              & \begin{tabular}[c]{@{}c@{}}Not \\ Smiling\end{tabular}             & 15.86          & 84.14                                                  & \begin{tabular}[c]{@{}c@{}}Not Bushy \\ Eyebrows\end{tabular}      & 42.23                                                     & 57.77                                                         & \begin{tabular}[c]{@{}c@{}}Not Pale \\ Skin\end{tabular}           & 28.75          & 71.25                                                    \\ \cline{3-12} 
                              &                                                                                  & \multirow{2}{*}{\textbf{\begin{tabular}[c]{@{}c@{}}After \\ Data Fine-tuning\end{tabular}}}  & Smiling                                                            & \textbf{73.26} & 26.74                                                  & \begin{tabular}[c]{@{}c@{}}Bushy \\ Eyebrows\end{tabular}          & \textbf{79.19}                                            & 20.81                                                         & Pale Skin                                                          & \textbf{76.57} & 23.43                                                    \\ \cline{4-12} 
                              &                                                                                  &                                                                                              & \begin{tabular}[c]{@{}c@{}}Not \\ Smiling\end{tabular}             & 10.76          & \textbf{89.24}                                         & \begin{tabular}[c]{@{}c@{}}Not Bushy \\ Eyebrows\end{tabular}      & 41.05                                                     & \textbf{58.95}                                                & \begin{tabular}[c]{@{}c@{}}Not Pale \\ Skin\end{tabular}           & 26.88          & \textbf{73.12}                                           \\ \hline
\end{tabular}
\end{table*}

\subsection{Implementation Details} 
The implementation details of training attribute classification model, perturbation learning, and model fine-tuning are discussed below.

\noindent \textbf{Training Attribute Classification Model:} To train attribute classification model pre-trained VGGFace$+$NNET is used. Two fully connected layers are used for training NNET of 512 dimensions. Each model is trained for 20 epochs with Adam optimizer, and learning rate is set to 0.005. 

\noindent \textbf{Perturbation Learning:} To learn the perturbation for a given dataset, learning rate is set to 0.001 and the batch size is 800. The number of iterations used for processing each batch is 16, and the number of epochs is 5.

\noindent \textbf{Model Fine-tuning:} To fine-tune the attribute classification model, Adam optimizer is used with learning rate set to 0.005. The model is trained for 20 epochs.

\begin{figure*}
\centering
\small
\includegraphics[scale = 0.41]{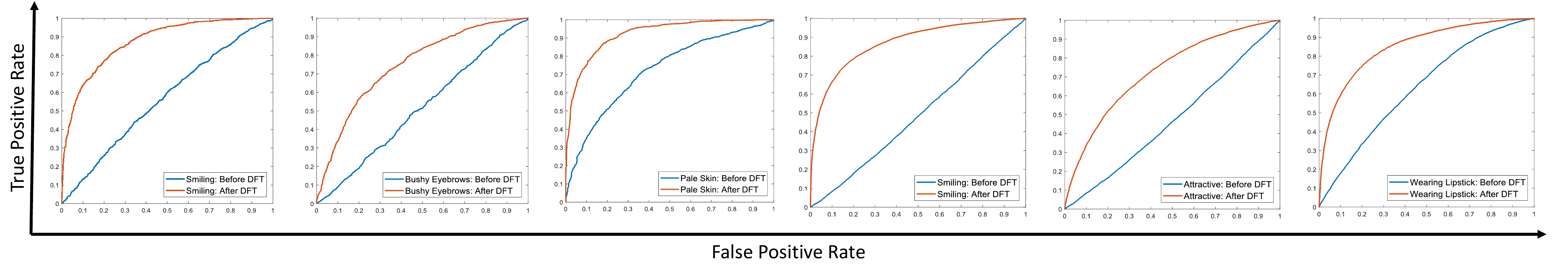}
\caption{ROC plots showing before and after data fine-tuning results of Black box Data Fine-tuning: Inter Dataset Experiment. First three ROC curves shows the result on the LFW dataset using a model trained on the CelebA dataset. Last three ROC curves shows the result on the CelebA dataset using a model trained on the LFW dataset (best viewed in color).}
\label{fig:ROC_Black_Box}
\end{figure*}

\section{Performance Evaluation}
The performance of the proposed algorithm is evaluated for \textit{Black Box Data Fine-tuning: Intra Dataset Experiment}, where the dataset used for data fine-tuning is same on which the pre-trained model is trained. On the other hand, in \textit{Black Box Data Fine-tuning: Inter Dataset Experiment}, the training data used to perform data fine-tuning is different from the training data used to train the pre-trained model. 

\subsection{Black Box Data Fine-tuning: Intra Dataset Experiment}
The proposed algorithm is evaluated on LFW, CelebA, and MUCT datasets for enhancing the performance of black box models. `Gender' is the common attribute among all three datasets. Table \ref{tab: Gender_Acc} shows the classification accuracy pertaining to before and after data fine-tuning for `Gender' attribute. For all three datasets, the classification accuracy improves by 1\% to 3\% using data fine-tuning. Specifically, the classification accuracy increases by 2.64\% for MUCT dataset whereas, for LFW dataset, the accuracy increases by 3.21\%. 

Three additional attributes, namely \textbf{LFW}-\{`Smiling', `Bushy Eyebrows', `Pale Skin'\}, \textbf{CelebA}-\{`Smiling', `Attractive', `Wearing Lipstick'\} are also evaluated. Table \ref{tab: Single_Attr_Class} shows the classification accuracy corresponding to these attributes. Similar to the results on `Gender' attribute, data fine-tuning leads to an overall increase in the classification accuracies of all the attributes for both the datasets. The classification accuracy of `Smiling' attribute increases by approximately 6\% for LFW dataset and 4\% for CelebA dataset. This shows the utility of data fine-tuning in enhancing the model's performance trained on the same dataset. 

Figure \ref{fig:Mis_Correct_Classified} shows some misclassified samples of LFW dataset corresponding to `Smiling', `Bushy Eyebrows', and `Pale Skin' attributes that are correctly classified after data fine-tuning. It is also observed that the visual appearance of the images is preserved. The score distribution of `Smiling' attribute, before and after data fine-tuning is shown in Figure \ref{fig: Score_Distr}. It is observed that the overlapping region between both the classes is reduced, and the confidence of predicting the true class scores is increased after data fine-tuning. The confusion matrix corresponding to the three attributes of the LFW dataset is shown in Table \ref{Conf_mat} which indicates that the True Positive Rate (TPR) and True Negative Rate (TNR) is improved for all three attributes. For instance, the TPR of `Smiling' attribute is increased by approximately 8\% and TNR is increased by approximately 5\% showcasing the efficacy of the proposed technique. 

\begin{table}
\scriptsize
\centering
\caption{Classification accuracy(\%) of Black box Data Fine-tuning: Inter Dataset experiment for `Gender' attribute on the MUCT, LFW, and CelebA datasets.}
\label{Gender:DD}
\begin{tabular}{|l|c|c|c|c|c|c|c|}
\hline
                                  & \multicolumn{1}{l|}{} & \multicolumn{6}{c|}{\textbf{Dataset used to train the model}}                                                                           \\ \hline
\multirow{2}{*}{}                 & \multirow{2}{*}{}     & \multicolumn{2}{c|}{\textbf{MUCT}} & \multicolumn{2}{c|}{\textbf{LFW}} & \multicolumn{2}{c|}{\textbf{CelebA}} \\ \cline{3-8} 
                                  &                       & Before       & After               & Before      & After               & Before        & After                \\ \hline
\multirow{3}{*}{\textbf{Dataset}} & \textbf{MUCT}         & -            & \textbf{-}          & 57.84       & \textbf{83.65}      & 80.27         & \textbf{92.84}       \\ \cline{2-8} 
                                  & \textbf{LFW}          & 63.09        & \textbf{80.45}      & -           & -                   & 56.01         & \textbf{86.33}       \\ \cline{2-8} 
                                  & \textbf{CelebA}       & 49.14        & \textbf{74.73}      & 67.53       & \textbf{76.59}      & -             & -                    \\ \hline
\end{tabular}
\end{table}

\begin{table}[t]
\scriptsize
\centering
\caption{Classification accuracy(\%) of Black box Data Fine-tuning: Inter Dataset experiment.}
\label{tab: BlackBox_Acc}
\begin{tabular}{|c|c|c|c|c|c|c|}
\hline
\multicolumn{1}{|l|}{}  & \multicolumn{6}{c|}{\textbf{Pre-trained Model trained on CelebA}}                                                                     \\ \hline
\multirow{3}{*}{LFW}    & \multicolumn{2}{c|}{\textbf{Smiling}} & \multicolumn{2}{c|}{\textbf{Bushy Eyebrows}} & \multicolumn{2}{c|}{\textbf{Pale Skin}}        \\ \cline{2-7} 
                        & Before        & After                 & Before            & After                    & Before             & After                     \\ \cline{2-7} 
                        & 55.29         & \textbf{78.61}        & 45.40             & \textbf{68.91}           & 56.62              & \textbf{84.21}            \\ \hline
\multicolumn{1}{|l|}{}  & \multicolumn{6}{c|}{\textbf{Pre-trained Model trained on LFW}}                                                                        \\ \hline
\multirow{3}{*}{CelebA} & \multicolumn{2}{c|}{\textbf{Smiling}} & \multicolumn{2}{c|}{\textbf{Attractive}}     & \multicolumn{2}{c|}{\textbf{Wearing Lipstick}} \\ \cline{2-7} 
                        & Before        & After                 & Before            & After                    & Before             & After                     \\ \cline{2-7} 
                        & 49.07         & \textbf{66.97}        & 49.71             & \textbf{66.60}           & 60.25              & \textbf{77.15}            \\ \hline
\end{tabular}
\end{table}

\begin{figure*}[t]
\centering
\includegraphics[scale = 0.57]{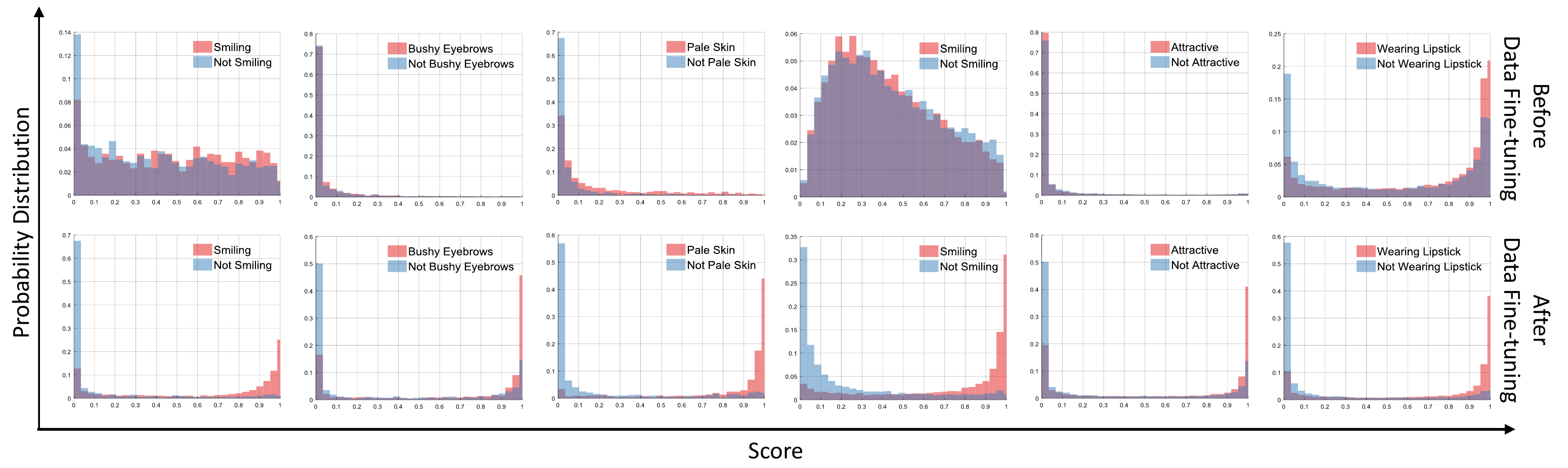}
\caption{Score distributions pertaining to before and after performing data fine-tuning. Top three graphs from the left represent the distribution ofthe  LFW dataset predicted using a model trained on the CelebA dataset. Bottom three graphs from the left represent its corresponding distribution after data fine-tuning. Similarly top three graphs from the right represent the score distribution on the CelebA dataset predicted using a model trained on the LFW dataset. Bottom three graphs from the right represent its corresponding distribution after data fine-tuning. (Best viewed in color).}
\label{fig: Score_Distr_Black_Box}
\end{figure*}

\subsection{Black box Data Fine-tuning: Inter Dataset Experiment}
This experiment is performed considering the real world scenario associated with Commercial off-the-shelf (COTS) systems where the training data distribution of the system is unknown to the user. The performance is evaluated for `Gender' attribute on all three datasets and the other three attributes used in Experiment 1 for LFW and CelebA datasets. 

Table \ref{Gender:DD} shows the classification accuracies for `Gender' attribute. It is observed that the classification accuracies increase by 12\% to 30\% on all three datasets. For other attributes on LFW and CelebA datasets, data fine-tuning is performed on the LFW dataset using a model trained on the attributes of the CelebA dataset and vice versa. Classification accuracies in Table \ref{tab: BlackBox_Acc} show the significant enhancement in the performance of the black box system using data fine-tuning. For instance, the accuracy on `Bushy Eyebrows' of the LFW dataset increases by approximately 23\%. Similarly, there is an improvement of 17\% on the attribute `Attractive' of the CelebA dataset. Figure \ref{fig:ROC_Black_Box} shows the ROC plots of all three attributes of LFW and CelebA datasets. The significant difference in the curves for all the attributes clearly demonstrates that the proposed algorithm is capable of improving the performance of the model with a large margin. Figure \ref{fig: Score_Distr_Black_Box} shows the score distribution before and after applying data fine-tuning. It is observed that before data fine-tuning, there is a huge overlap among the distributions of both the classes. For instance, the distribution of the attribute `Bushy Eyebrows' before perturbation for both the classes is on the same side resulting in higher misclassification rate. After data fine-tuning, the distribution of both the classes is well separated. This illustrates that data fine-tuning is able to shift the data corresponding to the model's unseen decision boundary.

\begin{figure}[t]
\centering
\includegraphics[scale = 0.26]{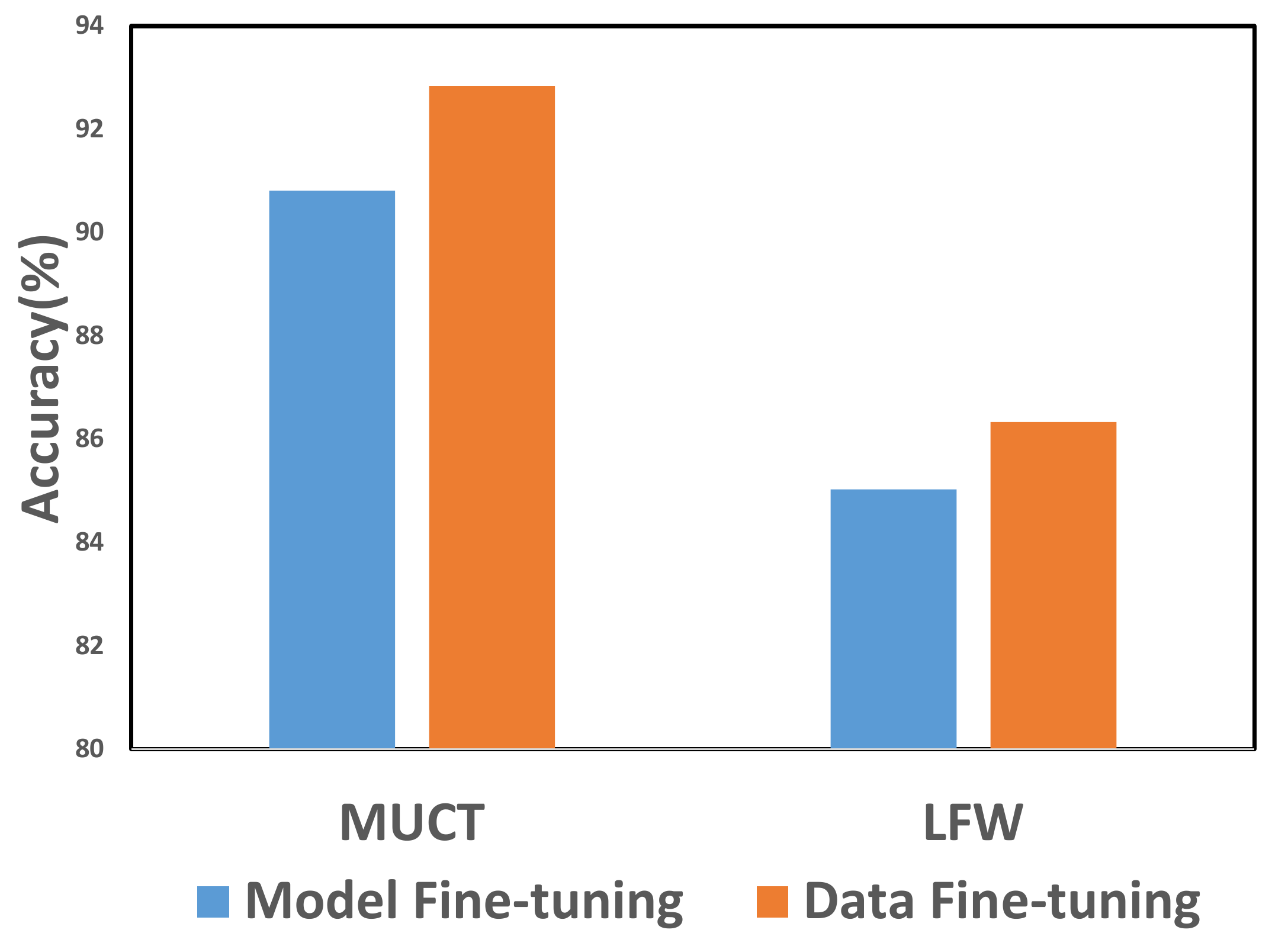}
\caption{Comparing Data Fine-tuning versus Model Fine-tuning for `Gender' attribute on the MUCT and LFW datasets using a model trained on the CelebA dataset.}
\label{fig:Bar_Gender}
\end{figure}

\begin{figure}
\centering
\includegraphics[scale = 0.29]{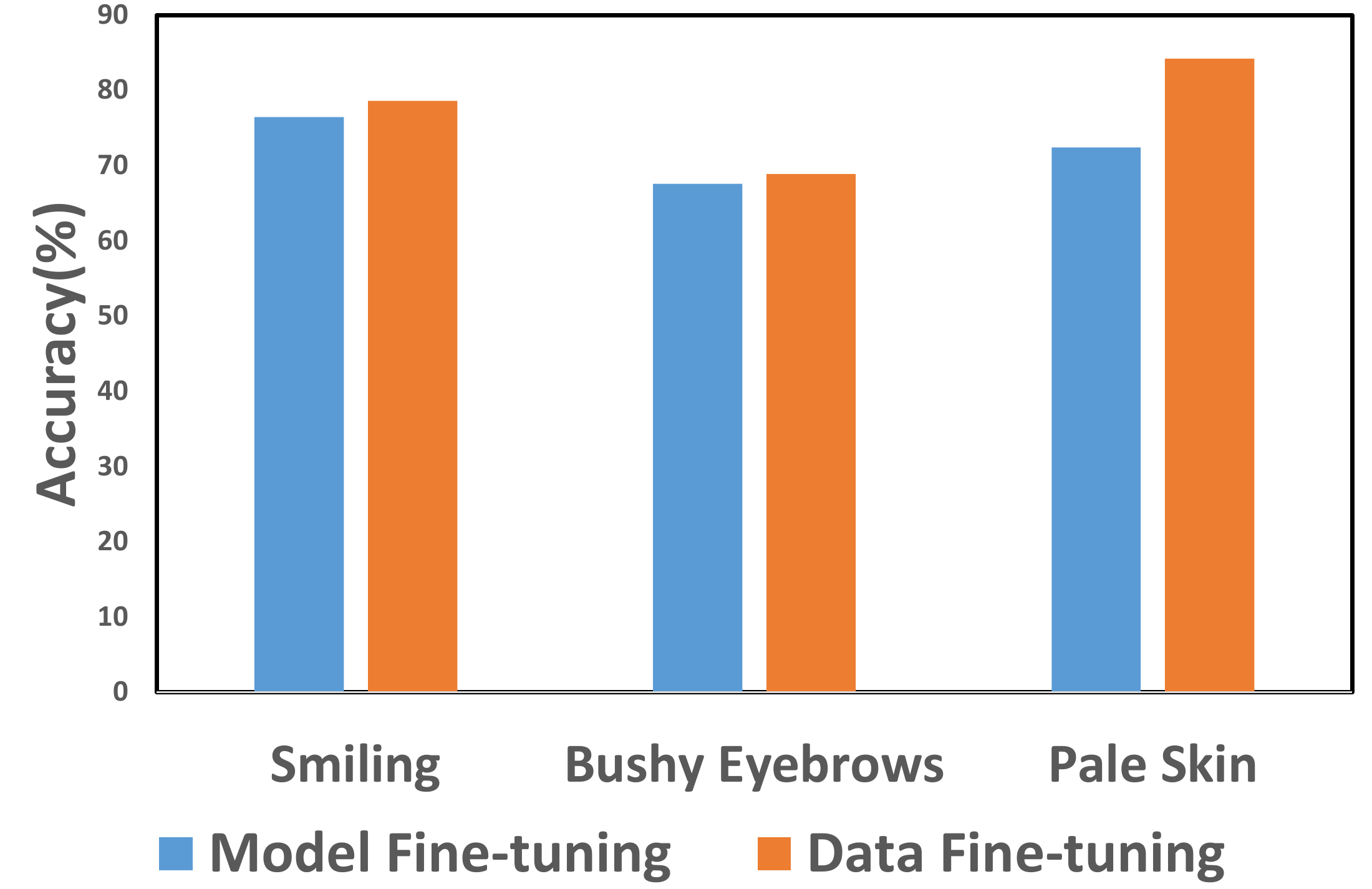}
\caption{Comparing the results of Data Fine-tuning versus Model Fine-tuning on the LFW dataset using a model trained on the CelebA dataset.}
\label{fig:Bar_Other}
\end{figure}

\subsection{Model Fine-tuning versus Data Fine-tuning}
This experiment is performed to compare the performance of model fine-tuning, where the model acts as a white box versus data fine-tuning where the model is a black box. For the experiments related to data fine-tuning, the procedure of `Black Box Data Fine-tuning: Inter Dataset Experiment' is followed. For model fine-tuning, the attribute classification model trained on the CelebA dataset is fine-tuned with MUCT and LFW dataset. Figure \ref{fig:Bar_Gender} shows the comparison of data fine-tuning with model fine-tuning for `Gender' attribute. In this experiment, the pre-trained model is trained on the CelebA dataset. On comparing the results on MUCT and LFW datasets, it is observed that data fine-tuning performs better than model fine-tuning for both the datasets. Experimental results obtained with other three attributes are shown in Figure \ref{fig:Bar_Other}, which also indicate that data fine-tuning outperforms model fine-tuning. Experiments are also performed by combining model fine-tuning with data fine-tuning. For this purpose, an iterative approach is followed, where data fine-tuning and model fine-tuning are performed iteratively. It is observed that the combination of model fine-tuning and data fine-tuning further enhances the results. However, such a combination is not useful for black-box systems where model fine-tuning is not possible.

\section{Conclusion}
Increasing demands of automated systems for face analysis has led to the development of several COTS systems. However, COTS systems are generally provided as black box systems and the model parameters are not available. In such scenarios, enhancing the performance of black-box systems is a challenging task. To address this situation, in this research a novel concept of data fine-tuning is proposed. Data fine-tuning refers to the process of adjusting the input data according to the behavior of the pre-trained model. The proposed data fine-tuning algorithm is designed using adversarial perturbation. Multiple experiments are performed to evaluate the performance of the proposed algorithm. It is observed that data fine-tuning enhances the performance of black box models. A comparison of data fine-tuning with model fine-tuning is also performed. We postulate that data fine-tuning can be an exciting alternative to model fine-tuning, particularly for black-box systems.

\section{Acknowledgements}
Vatsa and Singh are partially supported through Infosys Center for AI at IIIT Delhi, India. The authors acknowledge Shruti Nagpal for her constructive and useful feedback.

{\small
\bibliographystyle{aaai}
\bibliography{aaai_1}

\begin{thebibliography}{}
\bibitem[\protect\citeauthoryear{Hand, Castillo, and
  Chellappa}{2018}]{hand2018doing}
Hand, E.~M.; Castillo, C.~D.; and Chellappa, R.
\newblock 2018.
\newblock Doing the best we can with what we have: Multi-label balancing with
  selective learning for attribute prediction.
\newblock In {\em Association for the Advancement of Artificial Intelligence},
  6878--6885.
\end{thebibliography}
}
\end{document}